\documentclass[sigconf]{aamas}  

\usepackage{booktabs}

\usepackage{helvet}
\usepackage{courier}
\usepackage{url}
\usepackage{graphicx}
\usepackage{subfigure} 
\usepackage{dblfloatfix}

\usepackage{booktabs}

\usepackage{natbib}

\usepackage{comment}
\usepackage{amsmath}
\usepackage{amssymb}
\usepackage{siunitx}
\usepackage{wrapfig}
\usepackage{todonotes}
\usepackage{enumitem}
\usepackage{siunitx}
\usepackage{dsfont}
\usepackage{xcolor}
\usepackage{tablefootnote}
\usepackage{multirow}
\usepackage{hyperref}

\newcommand{\be}{\begin{equation}}
\newcommand{\ee}{\end{equation}}
\newcommand{\bea}{\begin{eqnarray}}
\newcommand{\eea}{\end{eqnarray}}
\newcommand{\beaa}{\begin{eqnarray*}}
\newcommand{\eeaa}{\end{eqnarray*}}

\DeclareMathAlphabet{\mathpzc}{OT1}{pzc}{m}{n}

\newcommand{\exs}[2]{{\mathbb E_{#1}}\left[ #2 \right]}

\usepackage{bm}

\usepackage{color}

\usepackage{url}

\usepackage{supertabular}

\numberwithin{equation}{section}

\numberwithin{thm}{section}

\DeclareMathAlphabet{\mathsfsl}{OT1}{cmss}{m}{sl}

\renewcommand{\phi}{\varphi}

\newcommand{\argmax}{\operatorname*{argmax}}

\newcommand{\Expect}{\operatorname{\mathbb{E}}}

\newcommand{\vct}[1]{\bm{#1}}

\usepackage{algorithm}
\usepackage{algorithmicx}
\usepackage[noend]{algpseudocode}
\usepackage{changepage}
\usepackage{bbm}

\algrenewcommand\algorithmicindent{1.0em}
\renewcommand{\algorithmiccomment}[1]{\hskip1em$\triangleright$ #1}

\setcopyright{ifaamas}  
\acmDOI{}  
\acmISBN{}  
\acmConference[AAMAS'18]{Proc.\@ of the 17th International Conference on Autonomous Agents and Multiagent Systems (AAMAS 2018)}{July 10--15, 2018}{Stockholm, Sweden}{M.~Dastani, G.~Sukthankar, E.~Andr\'{e}, S.~Koenig (eds.)}  
\acmYear{2018}  
\copyrightyear{2018}  
\acmPrice{}  

\usepackage{amsmath, amssymb, amsthm} 					
\usepackage{bm}											
\usepackage{graphicx}
\usepackage{caption}
\usepackage{dsfont}
\usepackage{booktabs}

\begin{document}

\title{
Learning with Opponent-Learning Awareness} 




%
\author{Jakob Foerster$^{\dagger, \ddagger}$}
\affiliation{%
 \institution{University of Oxford}
}

\author{Richard Y. Chen$^{\dagger}$}
\affiliation{%
 \institution{OpenAI}
}

\author{Maruan Al-Shedivat$^{ \ddagger}$}
\affiliation{%
 \institution{Carnegie Mellon University}
}

\author{Shimon Whiteson}
\affiliation{}
\affiliation{%
 \institution{University of Oxford}
}

\author{Pieter Abbeel$^{\ddagger}$}
\affiliation{}
\affiliation{%
 \institution{UC Berkeley}
}

\author{Igor Mordatch}
\affiliation{}
\affiliation{%
 \institution{OpenAI}
}
\renewcommand{\shortauthors}{J.\ Foerster et al.}

\thanks{$^\dagger$equal contribution, $^\ddagger$Work done at OpenAI, correspondence: jakob.foerster@cs.ox.ac.uk}

%

%
%
%
%
%
%
%

\begin{abstract}  
Multi-agent settings are quickly gathering importance in machine learning. This includes a plethora of recent work on deep multi-agent reinforcement learning, but also can be extended to hierarchical reinforcement learning, generative adversarial networks and decentralised optimization.
In all these settings the presence of multiple learning agents renders the training problem non-stationary and often leads to unstable training or undesired final results. We present \emph{Learning with Opponent-Learning Awareness} (LOLA), a method in which each agent shapes the anticipated learning of the other agents in the environment. The LOLA learning rule includes an additional term that accounts for the impact of one agent's policy on the anticipated parameter update of the other agents. Preliminary results show that the encounter of two LOLA agents leads to the emergence of tit-for-tat and therefore cooperation in the iterated prisoners' dilemma (IPD), while independent learning does not. In this domain, LOLA also receives higher payouts compared to a naive learner, and is robust against exploitation by higher order gradient-based methods. Applied to infinitely repeated matching pennies, LOLA agents converge to the Nash equilibrium. In a round robin tournament we show that LOLA agents can successfully shape the learning of a range of multi-agent learning algorithms from literature, resulting in the highest average returns on the IPD. We also show that the LOLA update rule can be efficiently calculated using an extension of the likelihood ratio policy gradient estimator, making the method suitable for model-free reinforcement learning. This method thus scales to large parameter and input spaces and nonlinear function approximators.  We also apply LOLA to a grid world task with an embedded social dilemma using deep recurrent policies and opponent modelling. Again, by explicitly considering the learning of the other agent, LOLA agents learn to cooperate out of self-interest. Our code is available at \href{https://www.github.com/alshedivat/lola}{github.com/alshedivat/lola}.

\end{abstract}

%


\maketitle


\section{Introduction}
\label{sec:introduction}

 Due to the advent of deep reinforcement learning (RL) methods that allow the study of many agents in rich environments, multi-agent RL has flourished in recent years. However, most of this recent work considers fully cooperative settings \citep{omidshafiei2017deep,foerster2017counterfactual,foerster2017stabilising} and emergent communication in particular \citep{das2017learning,mordatch2017emergence,lazaridou2016multi,foerster2016learning,sukhbaatar2016learning}. Considering future applications of multi-agent RL, such as self-driving cars, it is obvious that many of these will be only partially cooperative and contain elements of competition and conflict.

The human ability to maintain cooperation in a variety of complex social settings has been vital for the success of human societies.  Emergent reciprocity has been observed even in strongly adversarial settings such as wars \citep{axelrod2006evolution}, making it a quintessential and robust feature of human life.

In the future, artificial learning agents are likely to take an active part in human society, interacting both with other learning agents and humans in complex partially competitive settings. Failing to develop learning algorithms that lead to emergent reciprocity in these artificial agents would lead to disastrous outcomes.

How reciprocity can emerge among a group of learning, self-interested, reward maximizing RL agents is thus a question both of theoretical interest and of practical importance. Game theory has a long history of studying the learning outcomes in games that contain cooperative and competitive elements. In particular, the tension between cooperation and defection is commonly studied in the iterated prisoners' dilemma. In this game, selfish interests can lead to an outcome that is overall worse for all participants, while cooperation maximizes social welfare, one measure of which is the sum of rewards for all agents.

Interestingly, in the simple setting of an infinitely repeated prisoners' dilemma with discounting, randomly initialised RL agents pursuing independent gradient descent on the exact value function learn to defect with high probability. This shows that current state-of-the-art learning methods in deep multi-agent RL can lead to agents that fail to cooperate reliably even in simple social settings. 
One well-known shortcoming is that they fail to consider the learning process of the other agents and simply treat the other agent as a \emph{static} part of the environment~\citep{hernandez2017survey}. 

As a step towards reasoning over the learning behaviour of other agents in social settings, we propose \emph{Learning with Opponent-Learn-ing Awareness} (LOLA). The LOLA learning rule includes an additional term that accounts for the impact of one agent's policy on the learning step of the other agents. For convenience we use the word `opponents' to describe the other agents, even though the method is not limited to zero-sum games and can be applied in the general-sum setting. We show that this additional term, when applied by both agents, leads to emergent reciprocity and cooperation in the iterated prisoners' dilemma (IPD). Experimentally we also show that in the IPD, each agent is incentivised to switch from naive learning to LOLA, while there are no additional gains in attempting to exploit LOLA with higher order gradient terms. This suggests that within the space of local, gradient-based learning rules both agents using LOLA is a stable equilibrium. This is further supported by the good performance of the LOLA agent in a round-robin tournament, where it successfully manages to shape the learning of a number of multi-agent learning algorithms from literature. This leads to the overall highest average return on the IPD and good performance on Iterated Matching Pennies (IMP). 

We also present a version of LOLA adopted to the deep RL setting using likelihood ratio policy gradients, making LOLA scalable to settings with high dimensional input and parameter spaces. 

We evaluate the policy gradient version of LOLA on the IPD and iterated matching pennies (IMP), a simplified version of rock-paper-scissors. We show that LOLA leads to cooperation with high social welfare, while independent policy gradients, a standard multi-agent RL approach, does not. The policy gradient finding is consistent with prior work, e.g., \citet{sandholm1996multiagent}. 
We also extend LOLA to settings where the opponent policy is unknown and needs to be inferred from observations of the opponent's behaviour. 

Finally, we apply LOLA with and without opponent modelling to a grid-world task with an embedded underlying social dilemma. This task has temporally extended actions and therefore requires high dimensional recurrent policies  for agents to learn to reciprocate. Again, cooperation emerges in this task when using LOLA, even when the opponent's policy is unknown and needs to be estimated. 
\section{Related Work}
\label{sec:related_work}
The study of general-sum games has a long history in game theory and evolution. Many papers address the iterated prisoners' dilemma (IPD) in particular, including the seminal work on the topic by ~\citet{axelrod2006evolution}. This work popularised tit-for-tat (TFT), a strategy in which an agent cooperates on the first move and then copies the opponent's most recent move, as an effective and simple strategy. 

A  number of methods in  multi-agent RL aim to achieve convergence in self-play and rationality in sequential, general sum games. Seminal work includes the family of WoLF algorithms~\citep{bowling2002multiagent}, which uses different learning rates depending on whether an agent is winning or losing, joint-action-learners (JAL), and AWESOME~\citep{conitzer2007awesome}. 
Unlike LOLA, these algorithms typically have well understood convergence behaviour given an appropriate set of constraints.
However, none  of these algorithm have the ability to shape the learning behaviour of the opponents in order to obtain higher payouts at convergence. AWESOME aims to learn the equilibria of the one-shot game, a subset of the equilibria of the iterated game.

Detailed studies have analysed the dynamics of JALs in general sum settings: This includes work by~\citet{uther1997adversarial} in zero-sum settings and 
by~\citet{claus1998dynamics} in cooperative settings.
\citet{sandholm1996multiagent} study the dynamics of independent Q-learning in the IPD under a range of different exploration schedules and function approximators.
\citet{wunder2010classes} and \citet{zinkevich2006cyclic}  explicitly study the convergence dynamics and equilibria of learning in iterated games. Unlike LOLA, these papers do not propose novel learning rules.

\citet{littman2001friend} propose a method that assumes each opponent either to be a friend, i.e., fully cooperative, or foe, i.e., fully adversarial. Instead, LOLA considers general sum games.


 By comparing a set of models with different history lengths, \citet{chakraborty2014multiagent} propose a method to learn a best response to memory bounded agents with fixed policies. In contrast, LOLA assumes learning agents, which effectively correspond to unbounded memory policies.

\citet{brafman03} introduce the solution concept of an efficient learning equilibrium (ELE), in which neither side is encouraged to deviate from the learning rule. The algorithm they propose applies to settings where all Nash equilibria can be computed and enumerated; LOLA does not require either of these assumptions. 

By contrast, most work in deep multi-agent RL focuses on fully cooperative or zero-sum settings, in which learning progress is easier to evaluate, ~\citep{omidshafiei2017deep,foerster2017counterfactual,foerster2017stabilising} and emergent communication in particular ~\citep{das2017learning,mordatch2017emergence,lazaridou2016multi,foerster2016learning,sukhbaatar2016learning}. 
As an exception, ~\citet{leibo2017multi} analyse the outcomes of independent learning in general sum settings using feedforward neural networks as policies.
 ~\citet{lowe2017multi} propose a centralised actor-critic architecture for efficient training in these general sum environments. However, none of these methods explicitly reasons about the learning behaviour of other agents.
 ~\citet{Lanctot17PSRO} generalise the ideas of game-theoretic best-response-style algorithms, such as NFSP~\citep{heinrich2016deep}. In contrast to LOLA, these best-response algorithms assume a given set of opponent policies, rather than attempting to shape the learning of the other agents.

The problem setting and approach of \citet{lerer2017maintaining} is closest to ours.  They directly generalise tit-for-tat to complex environments using deep RL. The authors explicitly train a fully cooperative and a defecting policy for both agents and then construct a tit-for-tat policy that switches between these two in order to encourage the opponent to cooperate. Similar in spirit to this work, \citet{munoz08} propose a Nash equilibrium algorithm for repeated stochastic games that attempts to find the egalitarian equilibrium by switching between competitive and cooperative strategies. A similar idea underlies M-Qubed, \citep{crandall2011learning}, which balances best-response, cautious and optimistic learning biases. 

Reciprocity and cooperation are not emergent properties of the learning rules in these algorithms but rather directly coded into the algorithm via heuristics, limiting their generality.

 Our work also relates to opponent modelling, such as fictitious play ~\citep{brown1951iterative} and action-sequence prediction~\citep{ mealing2015opponent, rabinowitz2018machine}. \citet{mealing2013opponent} also 
 propose a method that finds a policy based on predicting the future action of a memory bounded opponent. Furthermore, \citet{hernandez2017learning} directly model the distribution over opponents. While these methods model the opponent strategy, or distribution thereof, and use look-ahead to find optimal response policies, they do not address the learning dynamics of opponents.
For further details we refer the reader to excellent reviews on the subject~\citep{hernandez2017survey, busoniu2008comprehensive}.

By contrast, \citet{zhang2010multi} carry out policy prediction under one-step learning dynamics. However, the opponents' policy updates are assumed to be given and only used to learn a best response to the anticipated updated parameters. By contrast,  a LOLA agent directly shapes the policy updates of all opponents in order to maximise its own reward. Differentiating through the opponent's learning step, which is unique to LOLA, is crucial for the emergence of tit-for-tat and reciprocity. To the best of our knowledge, LOLA is the first method that aims to shape the learning of other agents in a multi-agent RL setting.

With LOLA, each agent differentiates through the opponents' policy update. Similar ideas were proposed by \citet{metz2016unrolled}, whose training method for generative adversarial networks  differentiates through multiple update steps of the opponent. Their method relies on an end-to-end differentiable loss function, and thus does not work in the general RL setting. However, the overall results are similar: differentiating through the opponent's learning process stabilises the training outcome in a zero sum setting.

Outside of purely computational studies the emergence of cooperation and defection in RL settings has also been studied and compared to human data ~\cite{kleiman2016coordinate}.

\begin{figure*}[t]
  \includegraphics[width=0.95\textwidth]{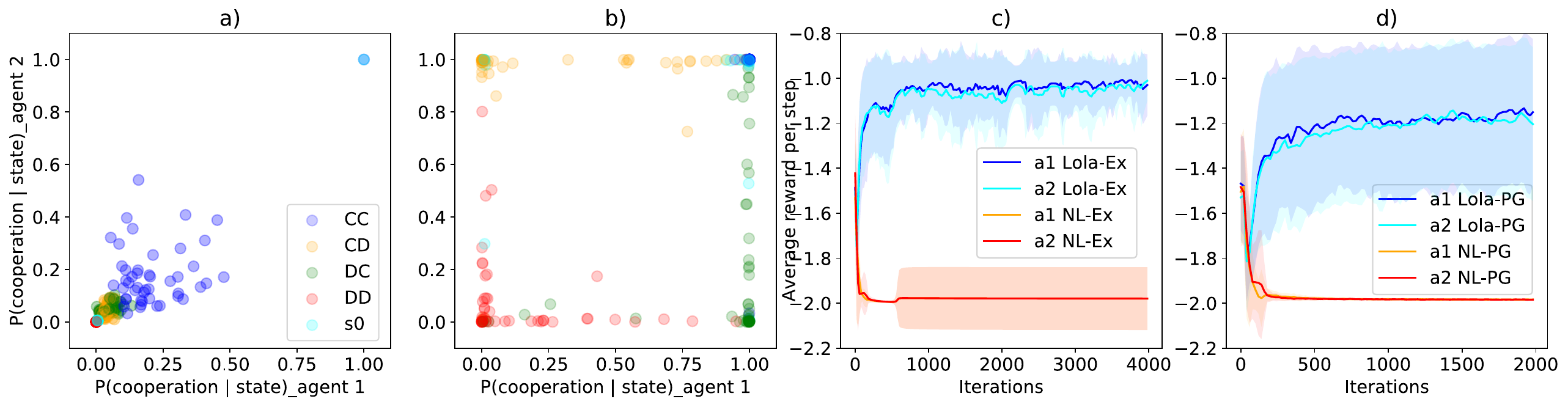}
  \vspace{-3ex}
\caption{ a) shows the probability of cooperation in the iterated prisoners dilemma (IPD) at the end of 50 training runs for both agents as a function of state under naive learning (NL-Ex) and b) displays the results for LOLA-Ex when using the exact gradients of the value function. c) shows the normalised discounted return for both agents in NL-Ex vs. NL-Ex and LOLA-Ex vs. LOLA-Ex, with the exact gradient. d) plots the normalised discounted return for both agents in NL-PG vs. NL-PG and LOLA-PG vs. LOLA-PG, with policy gradient approximation. We see that NL-Ex leads to DD, resulting in an average reward of ca. $-2$. In contrast, the LOLA-Ex agents play tit-for-tat in b): When in the last move agent 1 defected and agent 2 cooperated (DC, green points), most likely in the next move agent 1 will cooperate and agent 2 will defect, indicated by a concentration of the green points in the bottom right corner. Similarly, the yellow points (CD), are concentrated in the top left corner. While the results for the NL-PG and LOLA-PG with policy gradient approximation are more noisy, they are qualitatively similar. Best viewed in colour.}
\label{fig:fig_IPD}
\end{figure*}
\section{Notation}
\label{sec:background}

Our work assumes a multi-agent task that is commonly described as a stochastic game $G$, specified by a tuple $G = {\langle}S, U, P, r, Z, O, n, \gamma{\rangle}$. Here $n$ agents, $a \in A \equiv \{1,...,n\}$, choose actions, $u^a \in U$, and $s \in S$ is the state of the environment. The joint action $\mathbf{u} \in \mathbf{U} \equiv U^{n}$ leads to a state transition based on the transition function $P(s'|s,\mathbf{u}): S \times \mathbf{U} \times S \rightarrow [0,1]$. The reward functions $r^a(s,\mathbf{u}): S \times \mathbf{U} \rightarrow \mathbb{R}$ specify the rewards and $\gamma \in [0,1)$ is the discount factor.

We further define the discounted future return from time $t$ onward as $R^a_t = \sum_{l=0}^\infty \gamma^l r^a_{t+l}$ for each agent, $a$. In a naive approach, each agent maximises its total discounted return in expectation separately. This can be done with policy gradient methods \citep{sutton1999policy} such as REINFORCE~\citep{williams1992simple}. Policy gradient methods update an agent's policy, parameterised by $\vct{\theta}^a$, by performing gradient ascent on an estimate of the expected discounted total reward $\exs{}{R^a_0}$. 

By convention, bold lowercase letters denote column vectors.

\section{Methods}
\label{sec:method}
In this section, we review the naive learner's strategy and introduce the LOLA learning rule. We first derive the update rules when agents have access to exact gradients and Hessians of their expected discounted future return in Sections~\ref{sec:nl} and~\ref{sec:lola}. In Section~\ref{sec:pg}, we derive the learning rules purely based on policy gradients, thus removing access to exact gradients and Hessians. This renders LOLA suitable for deep RL. However, we still assume agents have access to opponents' policy parameters in policy gradient-based LOLA. Next, in Section~\ref{sec:om}, we incorporate opponent modeling into the LOLA learning rule, such that each LOLA agent only infers the opponent's policy parameter from experience. Finally, we discuss higher order LOLA in Section~\ref{sec:higher-oder-lola}.

For simplicity, here we assume the number of agents is $n=2$ and display the update rules for agent~1 only. The same derivation holds for arbitrary numbers of agents.
\subsection{Naive Learner} \label{sec:nl}
Suppose each agent's policy $\pi^a$ is parameterised by $\vct{\theta}^a$ and $V^{a}(\vct{\theta}^1, \vct{\theta}^2)$ is the expected total discounted return for agent $a$ as a function of both agents' policy parameters $(\vct{\theta}^1, \vct{\theta}^2)$. A Naive Learner (NL) iteratively optimises for its own expected total discounted return, such that at the $i$th iteration, $\vct{\theta}^a_i$ is updated to $\vct{\theta}^a_{i+1}$ according to
\begin{align*}
    \vct{\theta}^1_{i+1} & = \argmax\nolimits_{\vct{\theta}^1} V^1(\vct{\theta}^1, \vct{\theta}^2_i)\\
    \vct{\theta}^{2}_{i+1} & = \argmax\nolimits_{\vct{\theta}^2} V^2(\vct{\theta}^1_i, \vct{\theta}^2).
\end{align*}
In the reinforcement learning setting, agents do not have access to $\{V^1, V^2\}$ over all parameter values. Instead, we assume that agents only have access to the function values and gradients at $(\vct{\theta}^1_i, \vct{\theta}^2_i)$. Using this information the naive learners apply the gradient ascent update rule 
$\vct{f}^1_{\text{nl}}$:
\begin{align}
\vct{\theta}^1_{i+1} &= \vct{\theta}^1_{i} +   \vct{f}^1_{\text{nl}}(\vct{\theta}^1_{i}, \vct{\theta}^2_{i}), \nonumber \\
\vct{f}^1_{\text{nl}} &= \nabla_{\vct{\theta}^1_{i}} V^1(\vct{\theta}^1_{i}, \vct{\theta}^2_{i})   \cdot \delta, \label{eqn:f-nl}
\end{align}

where $\delta$ is the step size.

\subsection{Learning with Opponent Learning Awareness} \label{sec:lola}
A LOLA learner optimises its return under one step look-ahead of opponent learning: Instead of optimizing the expected return under the current parameters, $V^1(\vct{\theta}^1_{i}, \vct{\theta}^2_{i})$, a LOLA agent optimises $V^1(\vct{\theta}^1_{i}, \vct{\theta}^2_{i}  +\Delta \vct{\theta}^2_{i})$, which is the expected return after the opponent updates its policy with one naive learning step, $\Delta \vct{\theta}^2_{i}$. Going forward  we drop the subscript $i$ for clarity. Assuming small $\Delta \vct{\theta}^2$, a first-order Taylor expansion results in: 
\begin{equation} \label{eqn:lola-obj}
    V^1(\vct{\theta}^1, \vct{\theta}^2  +\Delta \vct{\theta}^2) \approx
    V^1(\vct{\theta}^1, \vct{\theta}^2)+ (\Delta \vct{\theta}^2)^T 
    \nabla_{\vct{\theta}^2}V^1(\vct{\theta}^1, \vct{\theta}^2).
\end{equation}
 
The LOLA objective \eqref{eqn:lola-obj} differs from prior work, e.g., \citet{zhang2010multi}, that predicts the opponent's policy parameter update and learns a best response. LOLA learners attempt to actively influence the opponent's future policy update, and explicitly differentiate through the $\Delta \vct{\theta}^2$ with respect to $\vct{\theta}^1$. Since LOLA focuses on this shaping of the learning direction of the opponent, the dependency of $\nabla_{\vct{\theta}^2} V^1(\vct{\theta}^1, \vct{\theta}^2)$ on $\vct{\theta}^1$ is dropped during the backward pass. Investigation of how differentiating through this term would affect the learning outcomes is left for future work. 

By substituting the opponent's naive learning step:
\begin{equation} \label{eqn:opponent-naive}
    \Delta \vct{\theta}^2 = \nabla_{\vct{\theta}^2} V^2(\vct{\theta}^1, \vct{\theta}^2) \cdot \eta
\end{equation}
into \eqref{eqn:lola-obj} and taking the derivative of \eqref{eqn:lola-obj} with respect to $\vct{\theta}^1$, we obtain our LOLA learning rule:
\begin{equation*}
\vct{\theta}^1_{i+1} = \vct{\theta}^1_{i}  + \vct{f}^1_{\text{lola}}(\vct{\theta}^1_{i}, \vct{\theta}^2_{i}),
\end{equation*}
 which includes a second order correction term
\begin{align}
&\vct{f}^1_{\text{lola}}(\vct{\theta}^1, \vct{\theta}^2) = \nabla_{\vct{\theta}^1} V^1(\vct{\theta}^1, \vct{\theta}^2)  \cdot \delta \nonumber \\
 & + \left(\nabla_{\vct{\theta}^2} V^1(\vct{\theta}^1, \vct{\theta}^2) \right)^T  
  \nabla_{\vct{\theta}^1} \nabla_{\vct{\theta}^2} V^2(\vct{\theta}^1, \vct{\theta}^2)  \cdot \delta  \eta, \label{eqn:f-lola}
\end{align}
where the step sizes $\delta, \eta$ are for the first and second order updates. 
Exact LOLA and NL agents (LOLA-Ex and NL-Ex) have access to the gradients and Hessians of $\{V^1, V^2\}$ at the current policy parameters $(\vct{\theta}^1_i, \vct{\theta}^2_i)$ and can evaluate \eqref{eqn:f-nl} and \eqref{eqn:f-lola} exactly.

\subsection{Learning via Policy Gradient} \label{sec:pg}
When agents do not have access to exact gradients or Hessians, we derive the update rules $f_{\text{nl, pg}}$ and $f_{\text{lola, pg}}$ based on approximations of the derivatives in \eqref{eqn:f-nl} and~\eqref{eqn:f-lola}. 
Denote an episode of horizon $T$ as $\tau = (s_0, u^1_0, u^2_0, r^1_0, r^2_0, ..., r^1_{T}, r^2_{T})$ and its corresponding discounted return for agent $a$ at timestep $t$ as
$R^a_t(\tau) = \sum\nolimits_{l=t}^{T} \gamma^{l-t} r^a_{l}$. 
The expected episodic return conditioned on the agents' policies $(\pi^1, \pi^2)$, $\Expect R^1_0(\tau)$ and $\Expect R^2_0(\tau)$, approximate $V^1$ and $V^2$ respectively, as do the gradients and Hessians. 

$\nabla_{\vct{\theta}^1}  \Expect R^1_0(\tau)$ follows from the policy gradient derivation:
\begin{align*}
&\nabla_{\vct{\theta}^1}  \Expect R^1_0(\tau) = \Expect\big[R^1_0(\tau) \nabla_{\vct{\theta}^1}\log \pi^1(\tau)\big]  \\
 &= \Expect\left[\sum\nolimits_{t=0}^{T} \nabla_{\vct{\theta}^1}  \log \pi^1(u^1_t|s_t) \cdot \sum\nolimits_{l=t}^{T} \gamma^l r^1_{l} \right] \\
 &= \Expect\left[\sum\nolimits_{t=0}^{T} \nabla_{\vct{\theta}^1}  \log \pi^1(u^1_t|s_t)  \gamma^t \big(R^1_t(\tau) - b(s_t)\big)\right],
\end{align*}
where $b(s_t)$ is a baseline for variance reduction. Then the update rule $\vct{f}_{\text{nl, pg}}$ for the policy gradient-based naive learner (NL-PG) is
\begin{equation} \label{eqn:f-nl-pg}
\vct{f}^1_{\text{nl, pg}} = \nabla_{\vct{\theta}^1}  \Expect R^1_0(\tau) \cdot \delta.
\end{equation}
For the LOLA update, we derive the following estimator of the second-order term in \eqref{eqn:f-lola} based on policy gradients. The derivation (omitted) closely resembles the standard proof of the policy gradient theorem, exploiting the fact that agents sample actions independently. We further note that this second order term is exact in expectation:
\begin{align}
\nabla_{\vct{\theta}^1}& \nabla_{\vct{\theta}^2} \Expect R^2_0(\tau)  \nonumber \\ &= \Expect\left[R^2_0(\tau)\nabla_{\vct{\theta}^1}\log \pi^1(\tau) \big(\nabla_{\vct{\theta}^2}\log \pi^2(\tau)\big)^T \right] \nonumber  \\
 & = \Expect \left[\sum\nolimits_{t=0}^{T} \gamma^t r^2_t\cdot  \left(\sum\nolimits_{l=0}^t \nabla_{\vct{\theta}^1} \log \pi^1(u^1_l|s_l) \right)\right. \nonumber  \\
& \quad \left. \left( \sum\nolimits_{l=0}^t \nabla_{\vct{\theta}^2} \log \pi^2(u^2_l|s_l) \right)^T \right]. \label{eqn:second-order}
\end{align} 
The complete LOLA update using policy gradients (LOLA-PG) is
\begin{align}
& \vct{f}^1_{\text{lola, pg}}= \nabla_{\vct{\theta}^1}  \Expect R^1_0(\tau) \cdot \delta + \nonumber \\
& \big(\nabla_{\vct{\theta}^2} \Expect R^1_0(\tau) \big)^T \nabla_{\vct{\theta}^1} \nabla_{\vct{\theta}^2} \Expect R^2_0 (\tau) \cdot \delta \eta.  \label{eqn:f-lola-pg}
\end{align}

\subsection{LOLA with Opponent Modeling} \label{sec:om}
Both versions \eqref{eqn:f-lola} and \eqref{eqn:f-lola-pg} of LOLA learning assume that each agent has access to the exact parameters of the opponent. However, in adversarial settings the opponent's parameters are typically obscured and have to be inferred from the opponent's state-action trajectories. Our proposed opponent modeling is similar to behavioral cloning~\cite{ross2011no,bojarski2016end}. Instead of accessing agent $2$'s true policy parameters $\vct{\theta}^2$, agent $1$ models the opponent's behavior with $\hat{\vct{\theta}}^2$, where $\hat{\vct{\theta}}^2$ is estimated from agent $2$'s trajectories using maximum likelihood:
\begin{equation}
\hat{\vct{\theta}}^2 =\argmax_{\vct{\theta}^2} \sum_t \log{\pi_{\vct{\theta}^2}(u^2_t|s_t)}.
\end{equation}
Then, $\hat{\vct{\theta}}^2$ replaces $\vct{\theta}^2$ in the LOLA update rule, both for the exact version \eqref{eqn:f-lola} using the value function and the gradient based approximation \eqref{eqn:f-lola-pg}. We compare the performance of policy-gradient based LOLA agents \eqref{eqn:f-lola-pg} with and without opponent modeling in our experiments. In particular we can obtain $\hat{\vct{\theta}}^2$ using the past action-observation history. In our experiments we incrementally fit to the most recent data in order to address the non-stationarity of the opponent.

\begin{figure*}[t]
  \includegraphics[width=0.95\textwidth]{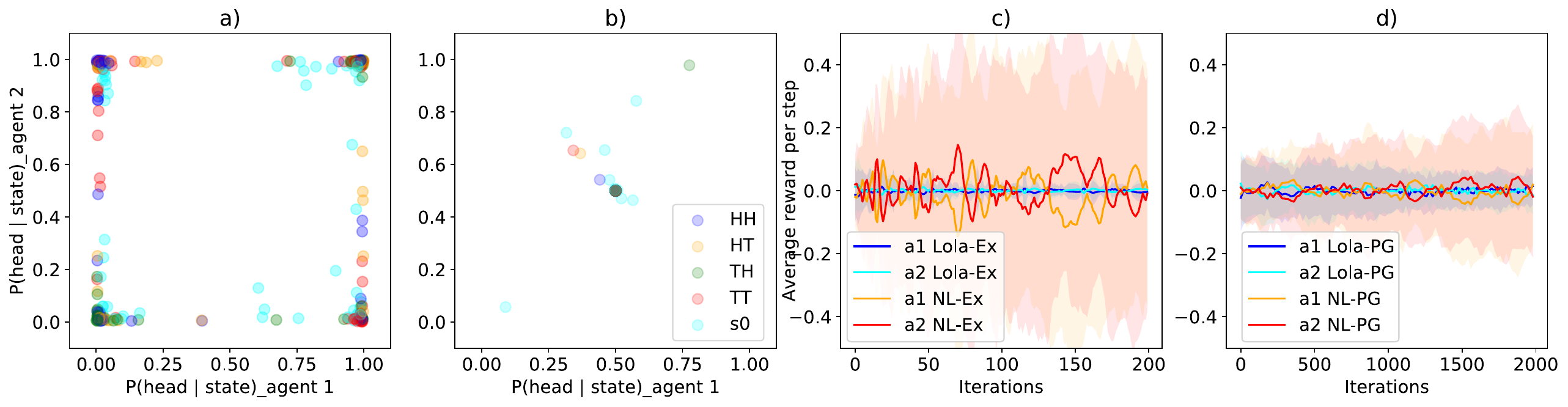}
  \vspace{-3ex}
\caption{ a) the probability of playing heads in the iterated matching pennies (IMP) at the end of 50 independent training runs for both agents as a function of state under naive learning NL-Ex. b) the results of LOLA-Ex when using the exact gradients of the value function. c) the normalised discounted return for both agents in NL-Ex vs. NL-Ex and LOLA-Ex vs. LOLA-Ex with exact gradient. d) the normalised discounted return for both agents in NL-PG vs. NL-PG and LOLA-PG vs. LOLA-PG with policy gradient approximation. We can see in a) that NL-Ex results in near deterministic strategies, indicated by the accumulation of points in the corners. These strategies are easily exploitable by other deterministic strategies leading to unstable training and high variance in the reward per step in c). In contrast, LOLA agents learn to play the only Nash strategy, $50\% /  50\%$, leading to low variance in the reward per step. One interpretation is that LOLA agents anticipate that exploiting a deviation from Nash increases their immediate return, but also renders them more exploitable by the opponent's next learning step. Best viewed in colour.}
\label{fig:fig_IMP}
\end{figure*}

\subsection{Higher-Order LOLA} \label{sec:higher-oder-lola}
By substituting the naive learning rule \eqref{eqn:opponent-naive} into the LOLA objective~\eqref{eqn:lola-obj}, the LOLA learning rule so far assumes that the opponent is a naive learner. We call this setting \emph{first-order LOLA}, which corresponds to the first-order learning rule of the opponent agent. However, we can also consider a higher-order LOLA agent that assumes the opponent applies a first-order LOLA learning rule, thus replacing \eqref{eqn:opponent-naive}. This leads to third-order derivatives in the learning rule. While the third-order terms are typically difficult to compute using policy gradient method, due to high variance, when the exact value function is available it is tractable. We examine the benefits of higher-order LOLA in our experiments.

\section{Experimental Setup}
\label{sec:setting}
In this section, we summarise the settings where we compare the learning behavior of NL and LOLA agents. The first setting (Sec.~\ref{subsec:iterated}) consists of two classic infinitely iterated games, the iterated prisoners dilemma (IPD), ~\cite{luce1957games} and iterated matching pennies (IMP)~\cite{lee1967application}. Each round in these two environments requires a single action from each agent. We can obtain the discounted future return of each player given both players' policies, which leads to exact policy updates for NL and LOLA agents. The second setting (Sec.~\ref{subsec:coin}), Coin Game, a more difficult two-player environment, where each round requires the agents to take a sequence of actions and exact discounted future reward can not be calculated. The policy of each player is parameterised with a deep recurrent neural network.

In the policy gradient experiments with LOLA, we assume off-line learning, i.e., agents play many (batch-size) parallel episodes using their latest policies. Policies remain unchanged within each episode, with learning happening between episodes. One setting in which this kind of offline learning naturally arises is when policies are trained on real-world data. For example, in the case of autonomous cars, the data from a fleet of cars is used to periodically train and dispatch new policies. 

\subsection{Iterated Games}
\label{subsec:iterated}
We first review the two iterated games, the IPD and IMP, and explain how we can model iterated games as a memory-$1$ two-agent MDP.

\begin{table}[h!]
\begin{center}
\begin{tabular}{c|c|c}
\hline
 & C & D \\
\hline
C & (-1, -1) & (-3, 0) \\
\hline
D & (0, -3) & (-2, -2)  \\
\hline
\end{tabular}
\end{center}
\caption{Payoff matrix of prisoners' dilemma.}
\label{tab:payoff}
\vspace{-3ex}
\end{table}

Table~\ref{tab:payoff} shows the per-step payoff matrix of the prisoners' dilemma. In a single-shot prisoners' dilemma, there is only one Nash equilibrium \cite{fudenberg1991game}, where both agents defect. In the infinitely iterated prisoners' dilemma, the folk theorem \citep{roger1991game} shows that there are infinitely many Nash equilibria. Two notable ones are the always defect strategy (DD), and tit-for-tat (TFT). In TFT each agent starts out with cooperation and then repeats the previous action of the opponent. The average returns per step in self-play are $-1$ and $-2$ for TFT and DD respectively. 

Matching pennies \cite{gibbons1992game} is a zero-sum game, with per-step payouts shown in Table~\ref{tab:pennies-payoff}. This game only has a single mixed strategy Nash equilibrium which is both players playing $50\% / 50\%$ heads / tails.  

\begin{table}[h!]
\begin{center}
\begin{tabular}{c|c|c}
\hline
 & Head & Tail \\
\hline
Head & (+1, -1) & (-1, +1) \\
\hline
Tail & (-1, +1) & (+1, -1)  \\
\hline
\end{tabular}
\end{center}
\caption{Payoff matrix of matching pennies.}
\label{tab:pennies-payoff}
\vspace{-3ex}
\end{table}

Agents in both the IPD and IMP can condition their actions on past history. Agents in an iterated game are endowed with a memory of length $K$, i.e., the agents act based on the results of the last $K$ rounds. Press and Dyson \cite{press2012iterated} proved that agents with a good memory-$1$ strategy can effectively force the iterated game to be played as memory-$1$. Thus, we consider memory-$1$ iterated games. 

We model the memory-$1$ IPD and IMP as a two-agent MRP, where the state at time $0$ is empty, denoted as $s_0$, and at time $t\geq1$ is the joint action from $t-1$:
    $s_t= (u^1_{t-1}, u^2_{t-1}) \quad \text{for $t>1$}$.

Each agent's policy is fully specified by $5$ probabilities. For agent $a$ in the case of the IPD, they are the probability of cooperation at game start $\pi^a(C|s_0)$, and the cooperation probabilities in the four memories: $\pi^a(C|CC)$, $\pi^a(C|CD)$, $\pi^a(C|DC)$, and $\pi^a(C|DD)$. By analytically solving the multi-agent MDP we can derive each agent's future discounted reward as an analytical function of the agents' policies  and calculate the exact policy update for both NL-Ex and LOLA-Ex agents.

We also organise a round-robin tournament where we compare LOLA-Ex to a number of state-of-the-art multi-agent learning algorithms, both on the  and IMP.

\begin{table}
\begin{center}
\begin{tabular}{|c|c|c|c|c|}
\hline
& \multicolumn{2}{c|}{IPD}   &  \multicolumn{2}{c|}{IMP} \\
& $\%$TFT & R(std) & $\%$Nash & R(std)\\
\hline
NL-Ex. &20.8 &-1.98(0.14) & 0.0& 0(0.37) \\
\hline
LOLA-Ex. &81.0 &-1.06(0.19) & 98.8 & 0(0.02) \\
\hline
NL-PG &20.0 & -1.98(0.00) & 13.2 & 0(0.19) \\
\hline
LOLA-PG &66.4  & -1.17(0.34) & 93.2 & 0(0.06) \\
\hline
\end{tabular}
\caption{We summarise results for NL vs. NL and LOLA vs. LOLA settings with either exact gradient evaluation (-Ex) or policy gradient approximation (-PG). Shown is the probability of agents playing TFT and Nash for the IPD and IMP respectively as well as the average reward per step, R, and standard deviation (std) at the end of training for 50 training runs.}
\label{tab:results}
\end{center}
\end{table}

\subsection{Coin Game}
\label{subsec:coin}
\begin{figure}[ht]
\includegraphics[width=0.95\columnwidth]{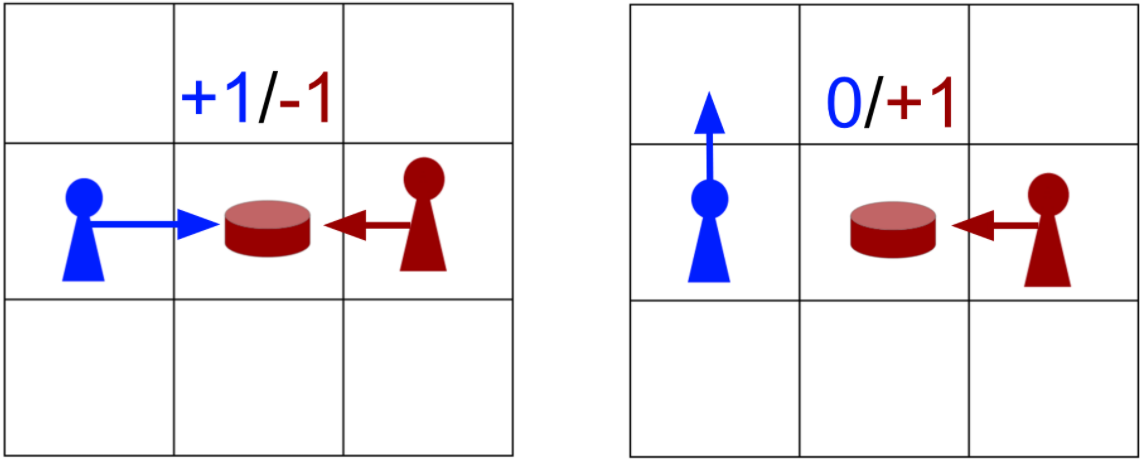}
 	\vspace{-3ex}
\caption{In the Coin Game, two agents, `red' and `blue', get 1 point for picking up any coin. However, the `red agent' loses 2 points when the `blue agent' picks up a red coin and vice versa. Effectively, this is a world with an embedded social dilemma where cooperation and defection are temporally extended.}
\label{fig:fig_coin_game}
\end{figure}

Next, we study LOLA in a more complex setting called Coin Game. This is a sequential game and the agents' policies are  parametrised as deep neural networks. Coin Game was first proposed by \citet{lerer2017maintaining} as a higher dimensional alternative to the IPD with multi-step actions. As shown in Figure~\ref{fig:fig_coin_game}, in this setting two agents, `red' and `blue', are tasked with collecting coins. 

The coins are either blue or red, and appear randomly on the grid-world. A new coin with random colour and random position appears after the last one is picked up. Agents pick up coins by moving onto the position where the coin is located. While every agent receives a point for picking up a coin of any colour, whenever an picks up a coin of different colour, the other agent loses 2 points. 

As a result, if both agents greedily pick up any coin available, they receive 0 points in expectation. Since the agents' policies are parameterised as a recurrent neural network, one cannot obtain the future discounted reward as a function of both agents' policies in closed form. Policy gradient-based learning is applied for both NL and LOLA agents in our experiments. We further include experiments of LOLA with opponent modelling (LOLA-OM) in order to examine the behavior of LOLA agents without access to the opponent's policy parameters.

\subsection{Training Details}
In policy gradient-based NL and LOLA settings, we train agents with an actor-critic method \citep{sutton1998reinforcement} and parameterise each agent with a policy actor and -critic for variance reduction during policy updates. 

During training, we use gradient descent with step size, $\delta$, of 0.005 for the actor, 1 for the critic, and the batch size 4000 for rollouts. The discount rate $\gamma$ is set to $0.96$ for the prisoners' dilemma and Coin Game and $0.9$ for matching pennies. The high value of $\gamma$ for Coin Game and the IPD was chosen in order to allow for long time horizons, which are known to be required for cooperation. We found that a lower $\gamma$ produced more stable learning on IMP.

For Coin Game the agent's policy architecture is a recurrent neural network with $32$ hidden units and 2 convolutional layers with $3\times3$ filters, stride $1$, and ReLU activation for input processing. The input is presented as a 4-channel grid, with 2 channels encoding the positions of the 2 agents and 2 channels for the red and blue coins respectively.

For the tournament, we use baseline algorithms and the corresponding hyperparameter values as provided in the literature~\citep{bowling2002multiagent}.
The tournament is played in a round-robin fashion between all pairs of agents for 1000 episodes, 200 steps each.

\begin{figure*}[t!]
 	\centering
 	\includegraphics[width=0.4\textwidth]{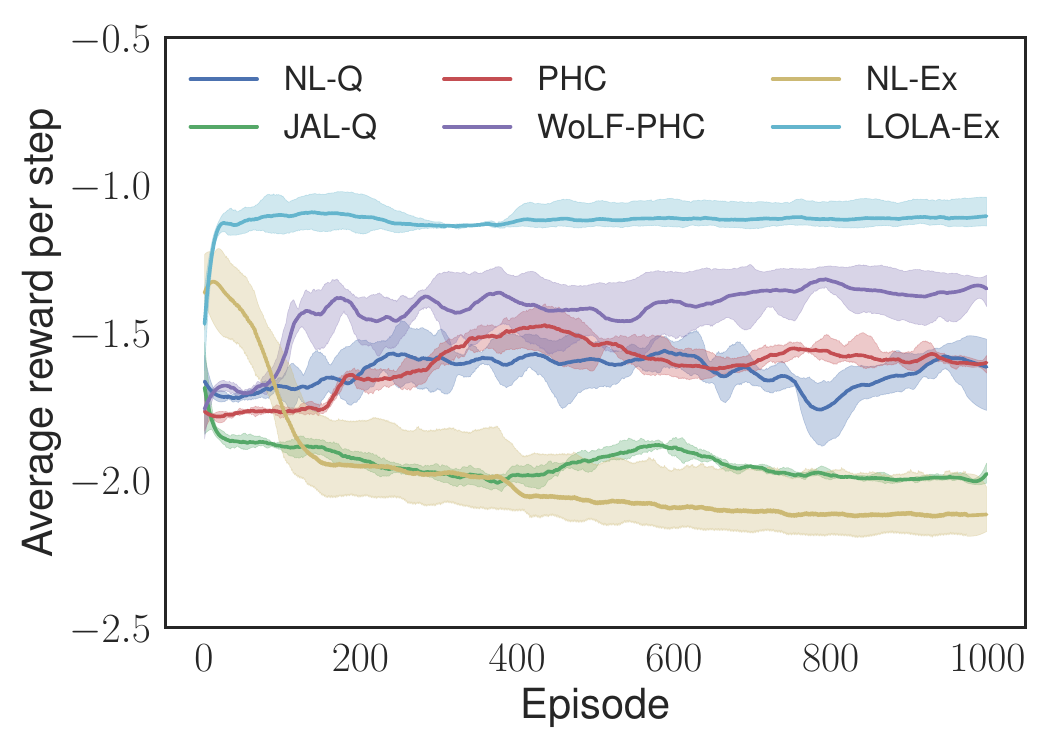}
 	\hfil
 	\includegraphics[width=0.4\textwidth]{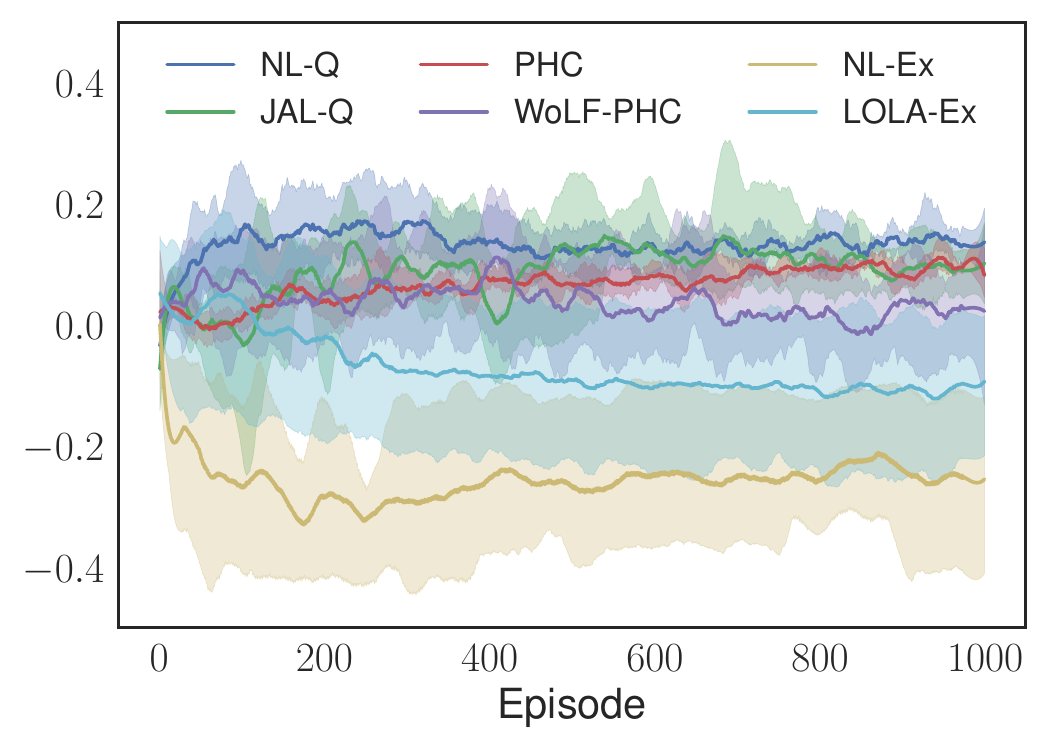}
 	\vspace{-3ex}
 	\caption{Normalised returns of a round-robin tournament on the IPD (left) and IMP (right). LOLA-Ex achieves the best performance in the IPD and is within error bars for IMP. Shading indicates a 95\% confidence interval for the mean. Baselines from~\citep{bowling2002multiagent}: naive Q-learner (NL-Q), joint-action Q-learner (JAL-Q), policy hill-climbing (PHC), and ``Win or Learn Fast'' (WoLF).}
 	\label{fig:tournament}
\end{figure*}

\section{Results}
In this section, we summarise the experimental results. We denote LOLA and naive agents with exact policy updates as LOLA-Ex and NL-Ex respectively. We abbreviate LOLA and native agents with policy updates with LOLA-PG and NL-PG. We aim to answer the following questions:
(1) How do pairs of LOLA-Ex agents behave in iterated games compared with pairs of NL-Ex agents? 
(2) Using policy gradient updates instead, how to LOLA-PG agents and NL-PG agents behave?
(3) How do LOLA-Ex agents fair in a round robin tournament involving a set of multi-agent learning algorithms from literature?
(4) Does the learning of LOLA-PG agents scale to high-dimensional settings where the agents' policies are parameterised by deep neural networks?
(5) Does LOLA-PG maintain its behavior when replacing access to the exact parameters of the opponent agent with opponent modeling?
(6) Can LOLA agents be exploited by using higher order gradients, i.e., does LOLA lead to an arms race of ever higher order corrections or is LOLA / LOLA stable?

We answer the first three questions in Sec.~\ref{subsec:iterated-results}, the next two questions in Sec.~\ref{subsec:coin-results}, and the last one in Sec.~\ref{subsec:higher-lola}.

\subsection{Iterated Games}
\label{subsec:iterated-results}
We first compare the behaviors of LOLA agents with NL agents, with either exact policy updates or policy gradient updates.

Figures~\ref{fig:fig_IPD}a and~\ref{fig:fig_IPD}b show the policy for both agents at the end of training under NL-Ex and LOLA-Ex when the agents have access to exact gradients and Hessians of $\{V^1, V^2\}$. Here we consider the settings of NL-Ex vs.\ NL-Ex and LOLA-Ex vs.\ LOLA-Ex. We study mixed learning of one LOLA-Ex agent vs.\ an NL-Ex agent in Section~\ref{subsec:higher-lola}. Under NL-Ex, the agents learn to defect in all states, indicated by the accumulation of points in the bottom left corner of the plot. However, under LOLA-Ex, in most cases the agents learn TFT. In particular agent 1 cooperates in the starting state $s_0$, $CC$ and $DC$, while agent 2 cooperates in $s_0$, $CC$ and $CD$. As a result, Figure~\ref{fig:fig_IPD}c) shows that the normalised discounted reward\footnote{We use following definition for the normalised discounted reward:  $ (1 - \gamma) \sum_{t=0}^T \gamma^t r_t $.} is close to $-1$ for LOLA-Ex vs.\ LOLA-Ex, corresponding to TFT, while NL-Ex vs.\ NL-Ex results in an normalised discounted reward of $-2$, corresponding to the fully defective ($DD$) equilibrium. Figure~\ref{fig:fig_IPD}d) shows the normalised discounted reward for NL-PG and LOLA-PG where agents learn via policy gradient. LOLA-PG also demonstrates cooperation while agents defect in NL-PG.

We conduct the same analysis for IMP in Figure~\ref{fig:fig_IMP}. In this game, under naive learning the agents' strategies fail to converge. In contrast, under LOLA the agents' policies converge to the only Nash equilibrium, playing $50\% / 50\%$ heads / tails. 

Table~\ref{tab:results} summarises the numerical results comparing LOLA with NL agents in both the exact and policy gradient settings in the two iterated game environments. In the IPD, LOLA agents learn policies consistent with TFT with a much higher probability and achieve higher normalised discounted rewards than NL ($-1.06$ vs $-1.98$). In IMP, LOLA agents converge to the Nash equilibrium more stably while NL agents do not. The difference in stability is illustrated by the high variance of the normalised discounted returns for NL agents compared to the low variance under LOLA ($0.37$ vs $0.02$).

In Figure~\ref{fig:tournament} we show the average normalised return of our LOLA-Ex agent against a set of learning algorithms from the literature. We find that LOLA-Ex receives the highest normalised return in the IPD, indicating that it successfully shapes the learning outcome of other algorithms in this general sum setting. 

In the IMP, LOLA-Ex achieves stable performance close to the middle of the distribution of results. 

\begin{figure*}[t!]
 	\centering
 	\includegraphics[width=0.45\textwidth]{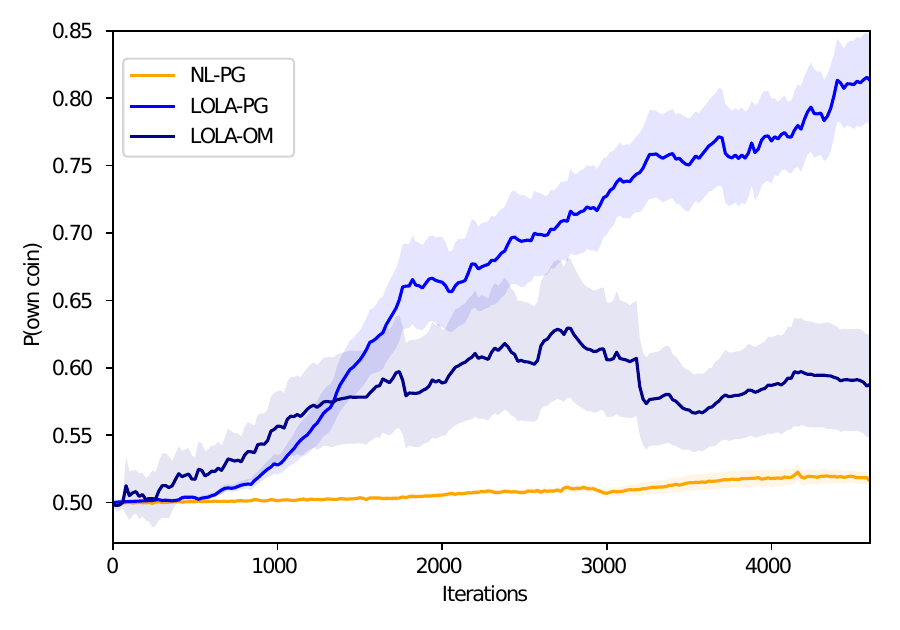}
 	\hfil
 	\includegraphics[width=0.45\textwidth]{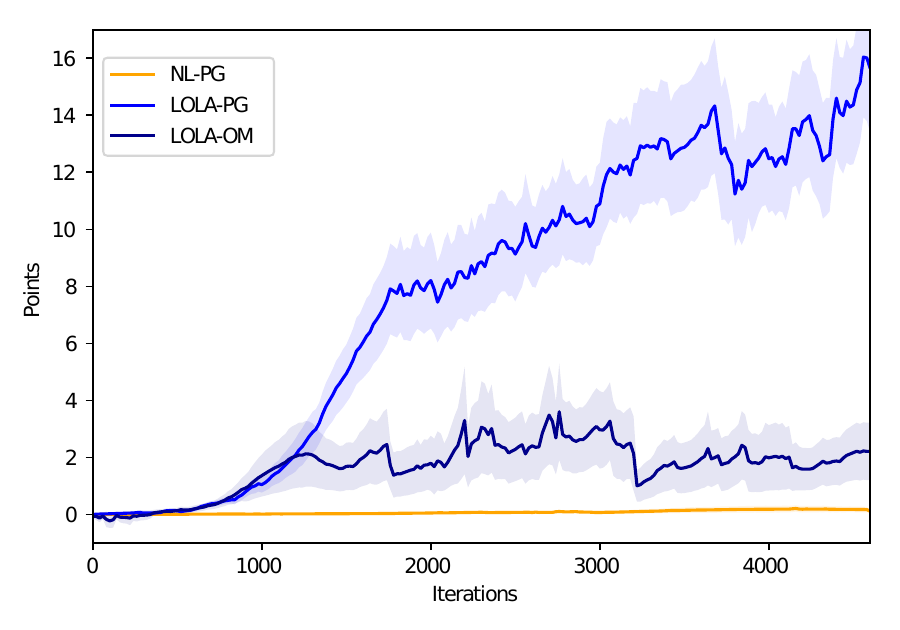}
 	\caption{The percentage of all picked up coins that match in colour (left) and the total points obtained per episode (right) for a pair of naive learners using policy gradient (NL-PG), LOLA-agents (LOLA-PG), and a pair of LOLA-agents with opponent modelling (LOLA-OM). Also shown is the standard error of the mean (shading), based on 10 training runs. While LOLA-PG and LOLA-OM agents learn to cooperate, LOLA-OM is less stable and obtains lower returns than LOLA-PG. Best viewed in colour.}
 	\label{fig:coin_game_results}
\end{figure*}

\subsection{Coin Game}
\label{subsec:coin-results}
We summarise our experimental results in the Coin Game environment. To examine the scalability of LOLA learning rules, we compare NL-PG vs.\ NL-PG to LOLA-PG vs.\ LOLA-PG. Figure~\ref{fig:coin_game_results} demonstrates that NL-PG agents collect coins indiscriminately, corresponding to defection. In contrast, LOLA-PG agents learn to pick up coins predominantly (around $80\%$) of their own colour, showing that the LOLA learning rule leads to cooperation in the Coin Game.

Removing the assumption that agents can access the exact parameters of opponents, we examine LOLA agents with opponent modeling (Section~\ref{sec:om}). Figure~\ref{fig:coin_game_results} demonstrates that without access to the opponent's policy parameters, LOLA agents with opponent modeling pick up coins of their own colour around $60\%$ of the time, inferior to the performance of LOLA-PG agents. We emphasise that with opponent modeling neither agent can recover the exact policy parameters of the opponent, since  there is a large amount of redundancy in the neural network parameters. For example, each agent could permute the weights of their fully connected layers. Opponent modeling introduces noise in the opponent agent's policy parameters, thus increasing the variance and bias of the gradients~\eqref{eqn:f-lola-pg} during policy updates, which leads to inferior performance of LOLA-OM vs.\ LOLA-PG in Figure~\ref{fig:coin_game_results}.

\subsection{Exploitability of LOLA}
\label{subsec:higher-lola}
In this section we address the exploitability of the LOLA learning rule. We consider the IPD, where one can calculate the exact value function of each agent given the policies. Thus, we can evaluate the higher-order LOLA terms. We pitch a NL-Ex or LOLA-Ex agent against NL-Ex, LOLA-Ex, and a 2nd-order LOLA agent. We compare the normalised discounted return of each agent in all settings and address the question of whether there is an arms race to incorporate ever higher orders of LOLA correction terms. 

Table~\ref{tab:higherorder} shows that a LOLA-Ex learner can achieve higher payouts against NL-Ex. Thus, there is an incentive for either agent to switch from naive learning to first order LOLA. Furthermore, two LOLA-Ex agents playing against each other both receive higher normalised discounted reward than a LOLA-Ex agent playing against a NL-Ex. This makes LOLA a dominant learning rule in the IPD compared to naive learning. 
We further find that 2nd-order LOLA provides no incremental gains when playing against a LOLA-Ex agent, leading to a reduction in payouts for both agents. These experiments were carried out with a $\delta$ of 0.5. While it is beyond the scope of this work to prove that LOLA vs LOLA is a dominant learning rule in the space of gradient-based rules, these initial results are encouraging.

\begin{table}[h!]
\begin{center}
\begin{tabular}{c|c|c|c}
\hline
 & NL-Ex & LOLA-Ex & 2nd-Order \\
\hline
NL-Ex & (-1.99, -1.99) & (-1.54, -1.28) & (-1.46, -1.46) \\
\hline
LOLA-Ex & (-1.28, -1.54) & (-1.04, -1.04) & (-1.14, -1.17) \\
\hline
\end{tabular}
\end{center}
\caption{Higher-order LOLA results on the IPD. A LOLA-Ex agent obtains higher normalised return compared to a NL-Ex agent. However in this setting there is no incremental gain from using higher-order LOLA in order to exploit another LOLA agent in the IPD. In fact both agents do worse with the 2nd order LOLA (incl. 3rd order corrections).}
\label{tab:higherorder}
\end{table}
 	\vspace{-6ex}

\section{Conclusions \& Future Work}
\label{sec:conclusion}
We  presented Learning with Opponent-Learning Awareness (LOLA), a learning method for multi-agent settings that considers the learning processes of other agents. We show that when both agents have access to exact value functions and apply the LOLA learning rule, cooperation emerges based on tit-for-tat in the infinitely repeated iterated prisoners' dilemma while independent naive learners defect. We also find that LOLA leads to stable learning of the Nash equilibrium in IMP. In our round-robin tournament against other multi-agent learning algorithms we show that exact LOLA agents achieve the highest average returns on the IPD and respectable performance on IMP. We also derive a policy gradient-based version of LOLA, applicable to a deep RL setting. Experiments on the IPD and IMP demonstrate similar learning behavior to the setting with exact value function. 

In addition, we scale the policy gradient-based version of LOLA to the Coin Game, a multi-step game that requires deep recurrent policies. LOLA agents learn to cooperate, as agents pick up coins of their own colour with high probability while naive learners pick up coins indiscriminately. We further remove agents' access to the opponent agents' policy parameters and replace with opponent modeling. While less reliable, LOLA agents with opponent modeling also learn to cooperate. 

We briefly address the exploitability of LOLA agents. Empirical results show that in the IPD both agents are incentivised to use LOLA, while higher order exploits show no further gain. 

In the future, we would like to continue to address the exploitability of LOLA, when adversarial agents explicitly aim to take advantage of a LOLA learner using global search methods rather than just gradient-based methods. Just as LOLA is a way to exploit a naive learner, there should be means of exploiting LOLA learners in turn, unless LOLA is itself an equilibrium learning strategy. 

\section*{Acknowledgements} 
This project has received funding from the European Research Council (ERC) under the European Union's Horizon 2020 research and innovation programme (grant agreement number 637713) and from the National Institutes of Health (grant agreement number R01GM114311).
It was also supported by the Oxford-Google DeepMind Graduate Scholarship and a generous equipment grant from NVIDIA.
We would like to thank Jascha Sohl-Dickstein, David Balduzzi, Karl Tuyls, Marc Lanctot, Michael Bowling, Ilya Sutskever, Bob McGrew, and Paul Cristiano for fruitful discussion. We would like to thank Michael Littman for providing feedback on an early version of the manuscript. We would like to thank our reviewers for constructive and thoughtful feedback.


\bibliographystyle{ACM-Reference-Format}  
\bibliography{ref}  

\end{document}


\title{Learning with Opponent-Learning Awareness \\\vspace{0.2cm}
	Supplementary Material}
	
\date{}
	
	\author{	}
	
	\maketitle
	
\appendix
\onecolumn
\section{Appendix}
\label{sec:appendix}
\subsection{Derivation of Second-Order derivative}
\label{sec:derive}
In this section, we derive the second order derivatives of LOLA in the policy gradient setting. Recall that an episode of horizon $T$ is 
\begin{equation*}
\tau = (s_0, u^1_0, u^2_0, r^1_0, r^2_0, ..., s_{T}, u^1_{T}, u^2_{T}, r^1_{T}, r^2_{T})
\end{equation*}
and the corresponding discounted return for agent $a$ at timestep $t$ is $R^a_t(\tau) = \sum\nolimits_{l=t}^{T} \gamma^{l-t} r^a_{l}$. We denote $\Expect_{\pi^1, \pi^2, \tau}$ as the expectation taken over both agents' policy and the episode $\tau$. Then,
\begin{align*}
\nabla_{\vct{\theta}^1}\nabla_{\vct{\theta}^2}\Expect_{\pi^1, \pi^2, \tau} R^1_0(\tau) &= \nabla_{\vct{\theta}^1}\nabla_{\vct{\theta}^2} \Expect_{\tau} \left[ R^1_0(\tau) \cdot \prod_{l=0}^{T} \pi^1(u^1_l| s_l, \vct{\theta}^1) \cdot \prod_{l=0}^{T} \pi^2(u^2_l | s_l, \vct{\theta}^2) \right] \\
& =   \Expect_{\tau} \left[ R^1_0(\tau) \cdot \left( \nabla_{\vct{\theta}^1} \big(\prod_{l=0}^{T}  \pi^1(u^1_l| s_l, \vct{\theta}^1) \big) \right) \left( \nabla_{\vct{\theta}^2} \big( \prod_{l=0}^{T} \pi^2(u^2_l | s_l, \vct{\theta}^2)\big) \right)^T \right] \\
& = \Expect_{\tau} \left[ R^1_0(\tau) \cdot \left( \frac{\nabla_{\vct{\theta}^1} \big( \prod_{l=0}^{T}  \pi^1(u^1_l| s_l, \vct{\theta}^1)\big)}{\prod_{l=0}^{T}  \pi^1(u^1_l| s_l, \vct{\theta}^1)} \right) \left( \frac{\nabla_{\vct{\theta}^2}\big(\prod_{l=0}^{T} \pi^2(u^2_l | s_l, \vct{\theta}^2)\big)}{\prod_{l=0}^{T} \pi^2(u^2_l | s_l, \vct{\theta}^2)} \right)^T  \right. \\
& \left. \quad \cdot \prod_{l=0}^{T}  \pi^1(u^1_l| s_l, \vct{\theta}^1)  \cdot  \prod_{l=0}^{T} \pi^2(u^2_l | s_l, \vct{\theta}^2)\right] \\
& = \Expect_{\pi^1, \pi^2, \tau} \left[ R^1_0(\tau) \cdot \left(\frac{\nabla_{\vct{\theta}^1} \big( \prod_{l=0}^{T}  \pi^1(u^1_l| s_l, \vct{\theta}^1)\big)}{\prod_{l=0}^{T}  \pi^1(u^1_l| s_l, \vct{\theta}^1)} \right) \left( \frac{\nabla_{\vct{\theta}^2}\big(\prod_{l=0}^{T} \pi^2(u^2_l | s_l, \vct{\theta}^2)\big)}{\prod_{l=0}^{T} \pi^2(u^2_l | s_l, \vct{\theta}^2)} \right)^T  \right]\\
& = \Expect_{\pi^1, \pi^2, \tau} \left[ R^1_0(\tau) \cdot \left(\nabla_{\vct{\theta}^1} \log\big( \prod_{l=0}^{T}  \pi^1(u^1_l| s_l, \vct{\theta}^1)\big) \right) \left( \nabla_{\vct{\theta}^2}\log\big(\prod_{l=0}^{T} \pi^2(u^2_l | s_l, \vct{\theta}^2)\big)\right)^T  \right] \\
& = \Expect_{\pi^1, \pi^2, \tau} \left[ R^1_0(\tau) \cdot \left(\sum\nolimits_{l=0}^{T} \nabla_{\vct{\theta}^1} \log \pi^1(u^1_l|s_l, \vct{\theta}^1) \right)  \left( \sum\nolimits_{l=0}^{T} \nabla_{\vct{\theta}^2} \log \pi^2(u^2_l|s_l, \vct{\theta}^2) \right)^T \right].
\end{align*}
The second equality is due to $\pi_l$ is only a function of $\vct{\theta}_l$. The third equality is multiply and divide the probability of the episode $\tau$. The fourth equality factors the probability of the episode $\tau$ into the expectation $\Expect_{\pi^1, \pi^2, \tau}$. The fifth and sixth equalities are standard policy gradient operations.

Similar derivations lead to the the following second order cross-term gradient for a single reward of agent $1$ at time $t$
\begin{align*}
\nabla_{\vct{\theta}^1}\nabla_{\vct{\theta}^2} \Expect_{\pi^1, \pi^2, \tau} r^1_t &= \Expect_{\pi^1, \pi^2, \tau} \left[r^1_t \cdot\left( \sum\nolimits_{l=0}^t \nabla_{\vct{\theta}^1} \log \pi^1(u^1_l|s_l, \vct{\theta}^1) \right) \left(\sum\nolimits_{l=0}^t \nabla_{\vct{\theta}^2} \log \pi^2(u^2_l|s_l, \vct{\theta}^2) \right)^T \right].
\end{align*}
Sum the rewards over $t$,
\begin{align*}
\nabla_{\vct{\theta}^1}\nabla_{\vct{\theta}^2} \Expect_{\pi^1, \pi^2, \tau} R^1_0(\tau) &= \Expect_{\pi^1, \pi^2, \tau} \left[\sum\nolimits_{t=0}^{T} \gamma^t r^1_t\cdot  \left( \sum\nolimits_{l=0}^t \nabla_{\vct{\theta}^1} \log \pi^1(u^1_l|s_l, \vct{\theta}^1)\right)
\left( \sum\nolimits_{l=0}^t \nabla_{\vct{\theta}^2} \log \pi^2(u^2_l|s_l, \vct{\theta}^2) \right)^T \right],
\end{align*}
which is the 2nd order term in the Methods Section.

\subsection{Derivation of the exact value function in the Iterated Prisoners' dilemma and Iterated Matching Pennies}
\label{sec:discount-rew}
In both IPD and IMP the action space consists of 2 discrete actions. 
The state consists of the union of the last action of both agents. As such there are a total of 5 possible states, 1 state being the initial state, $s_0$,  and the other 4 the 2 x 2 states depending on the last action taken.

As a consequence the policy of each agent can be represented by $5$ parameters,$\vct{\theta}^{a}$,the probabilities of taking action $0$ in each of these $5$ states. In the case of the IPD these parameters correspond to the probability of cooperation in $s_0$, CC, CD, DC and DD:
\begin{align*}
\pi^a(C|s_0) & = {\theta}^{a, 0}, \quad \pi^a(D|s_0)  = 1 - {\theta}^{a, 0}, \\
\pi^a(C|CC) &= {\theta}^{a, 1}, \quad \pi^a(D|CC) = 1- {\theta}^{a,1}, \\
\pi^a(C|CD) &= {\theta}^{a, 2}, \quad \pi^a(D|CD) = 1- {\theta}^{a,2}, \\
\pi^a(C|DC) &= {\theta}^{a, 3}, \quad \pi^a(D|DC) = 1- {\theta}^{a,3}, \\
\pi^a(C|DD) &= {\theta}^{a, 4}, \quad \pi^a(D|DD) = 1- {\theta}^{a,4}, \quad a \in \{1, 2\}.
\end{align*}
We denote $\vct{\theta}^a=({\theta}^{a,0}, {\theta}^{a,1}, {\theta}^{a,2}, {\theta}^{a,3}, {\theta}^{a,4})$. 

In these games the union of $\pi^1$ and $\pi^2$ induces a state transition function $P(s'|s) = P(\vct{u}|s)$.
Denote the distribution of $s_0$ as $\vct{p}_0$:
\begin{equation*}
\vct{p}_0 = \big({\theta}^{1,0}{\theta}^{2,0}, \ {\theta}^{1,0}(1-{\theta}^{2,0}),\ (1-{\theta}^{1,0}){\theta}^{2,0},\ (1-{\theta}^{1,0})(1-{\theta}^{2,0}) \big)^T,
\end{equation*}
the payout vector as 
\begin{equation*}
\vct{r}^1=(-1, -3, 0, -2 )^T \quad \text{and} \quad \vct{r}^2=(-1, 0, -3, -2)^T,
\end{equation*}
and the transition matrix is
\begin{equation*}
 \mtx{P} =\left[  
 \begin{matrix}
 \vct{\theta}^{1} \vct{\theta}^{2}, &  \vct{\theta}^{1} (\onevctt - \vct{\theta}^{2} ), & (\vct{\theta}^{1} - \onevctt) \vct{\theta}^{2}, & (\onevctt- \vct{\theta}^{1})(\onevctt - \vct{\theta}^{2} )
 \end{matrix}
 \right]
\end{equation*}
Then $V_1, V_2$ can be represented as
\begin{align*}
V^1(\vct{\vct{\theta}}^1, \vct{\theta}^2)& =  \vct{p}_0^T \big( \vct{r}^1 + \sum\nolimits_{t=1}^{\infty}\gamma^t \mtx{P}^t \vct{r}^1 \big)  \\
V^2(\vct{\theta}^1, \vct{\theta}^2)& = \vct{p}_0^T \big( \vct{r}^2 + \sum\nolimits_{t=1}^{\infty} \gamma^t \mtx{P}^t \vct{r}^2    \big).
\end{align*}
Since $\gamma < 1$ and $\mtx{P}$ is a stochastic matrix, the infinite sum converges and
\begin{align*}
V^1(\vct{\theta}^1, \vct{\theta}^2)& = \vct{p}_0^T\ \frac{\Id}{\Id - \gamma\mtx{P}}\ \vct{r}^1, \\
V^2(\vct{\theta}^1, \vct{\theta}^2)& = \vct{p}_0^T\ \frac{\Id}{\Id - \gamma\mtx{P}}\ \vct{r}^2,
\end{align*}
where $\Id$ is the identity matrix. 

An equivalent derivation holds for the Iterated Matching Pennies game with $\vct{r}^1=(-1, 1, 1, -1)^T$ and $\vct{r}^2 = - \vct{r}^1$.

\newpage
\subsection{Figures}
\label{sec:append_fig}
\begin{figure*}[ht]
\centering
	\subfigure[]{
	\includegraphics[width=1\textwidth]{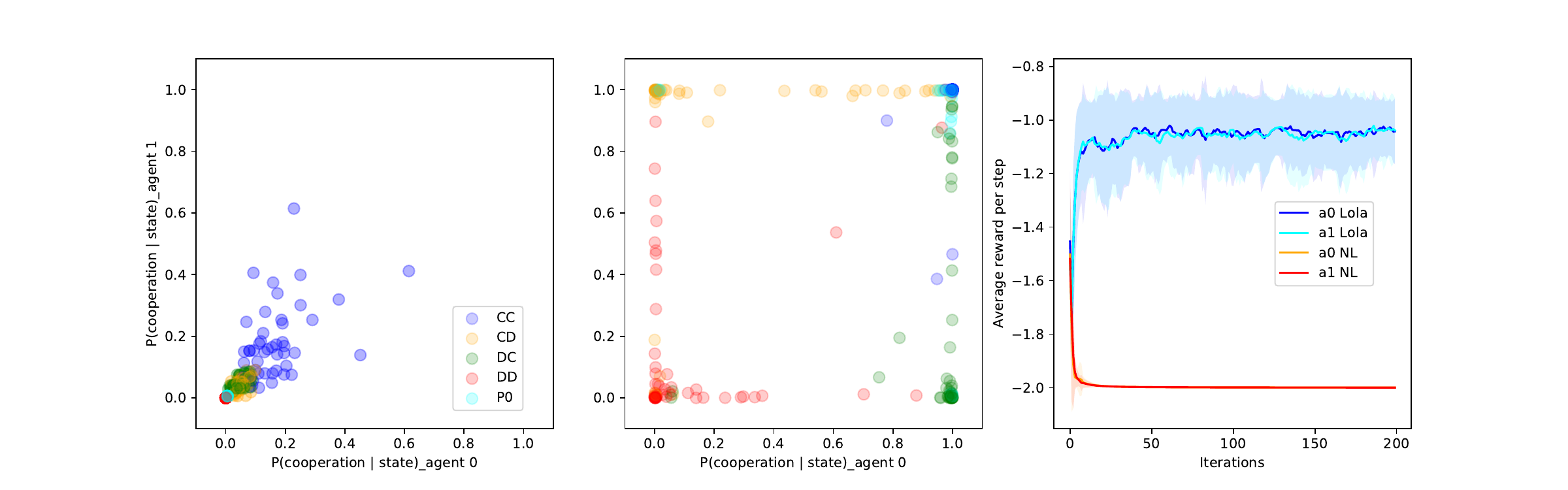}
	}
	\subfigure[]{
	\includegraphics[width=1\textwidth]{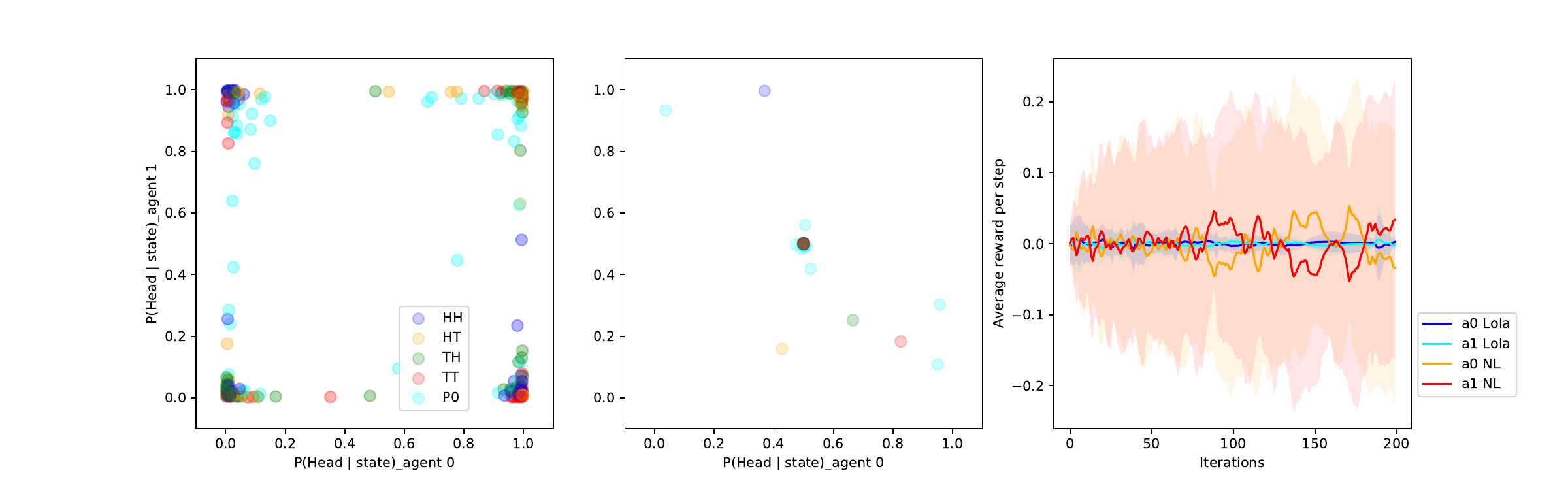}
	}
\label{fig:exact_V}
\caption{Shown is the probability of cooperation in the prisoners dilemma (a) and the probability of heads in the matching pennies game (b) at the end of 50 training runs for both agents as a function of state under naive learning (left) and LOLA (middle) when using the exact gradients of the value function. Also shown is the average return per step for naive and LOLA (right) }
\end{figure*}

\begin{figure*}[ht]
\centering
	\subfigure[]{
	\includegraphics[width=1\textwidth]{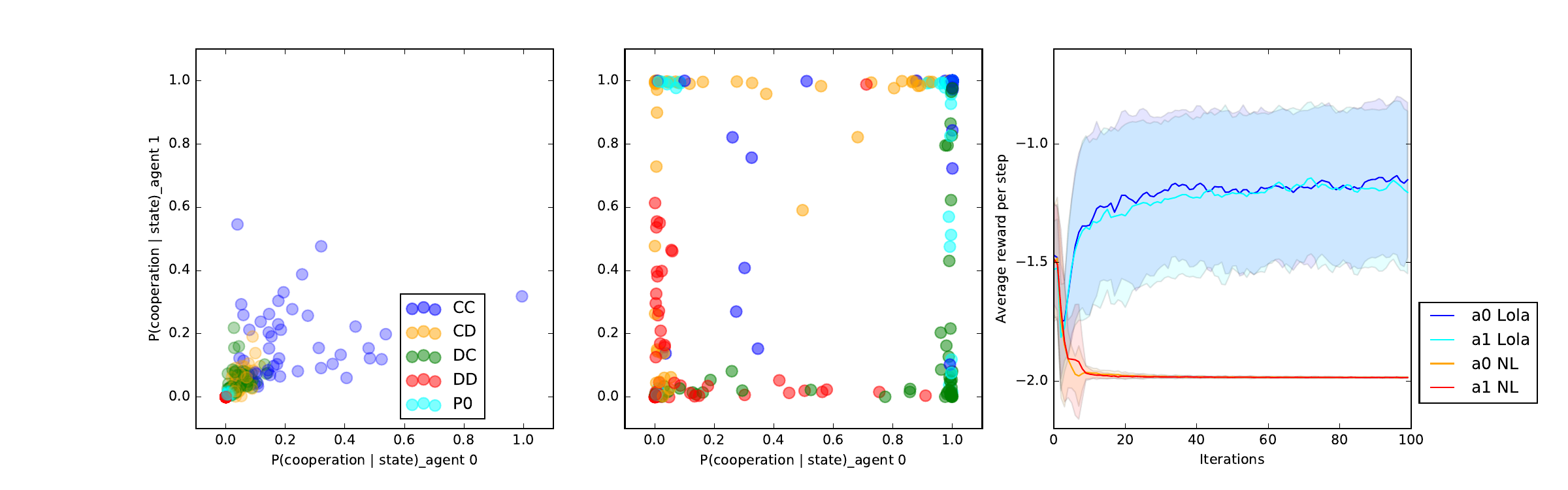}
	}
	\subfigure[]{
	\includegraphics[width=1\textwidth]{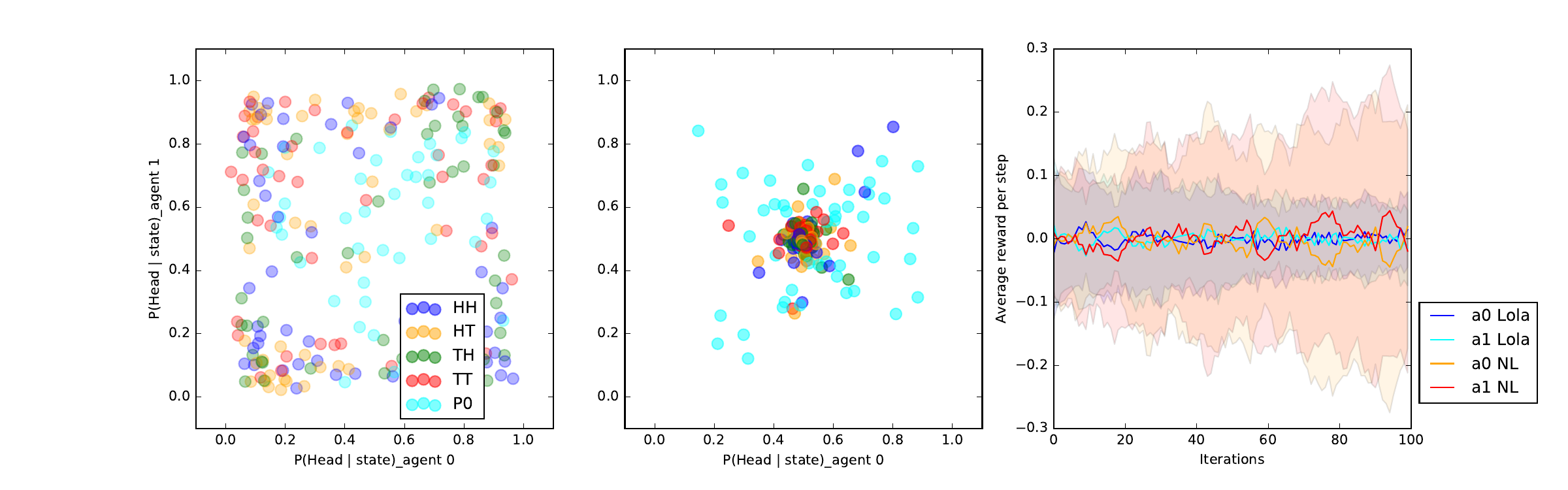}
	}
\label{fig:PG}
\caption{Same as Figure~\ref{fig:exact_V}, but using the policy gradient approximation for all terms. Clearly results are more noisy by qualitatively follow the results of the exact method.}
\end{figure*}